\documentclass[letterpaper, 10 pt, conference]{ieeeconf}
\IEEEoverridecommandlockouts
\usepackage{amsmath,amsfonts}
\usepackage[noend]{algorithmic}
\usepackage{array}
\usepackage[caption=false,font=normalsize,labelfont=sf,textfont=sf]{subfig}
\usepackage{textcomp}
\usepackage{stfloats}
\usepackage{url}
\usepackage{verbatim}
\usepackage{graphicx}
\usepackage{algorithm}   
\usepackage{tabularx}
\usepackage{setspace} 
\usepackage{subcaption}
\usepackage{caption}
\usepackage[top=57pt,bottom=43pt,left=48pt,right=48pt]{geometry}
\usepackage[dvipsnames]{xcolor}
\usepackage{colortbl}
\usepackage{booktabs}
\usepackage{multirow}
\usepackage{multicol}
\usepackage{ifthen}

\newif\ifredcolor
\redcolorfalse
\newcommand{\red}[1]{%
  \ifthenelse{\boolean{redcolor}}%
  {\textcolor{red}{#1}}%
  {#1}%
}
\newcommand{\sectionred}[1]{%
  \ifthenelse{\boolean{redcolor}}%
    {\section{\textcolor{red}{#1}}}%
    {\section{#1}}%
}

\let\labelindent\relax
\usepackage[inline]{enumitem}
\def\BibTeX{{\rm B\kern-.05em{\sc i\kern-.025em b}\kern-.08em
    T\kern-.1667em\lower.7ex\hbox{E}\kern-.125emX}}
\usepackage{balance}
\usepackage[backend=biber,
            hyperref=true,
            url=false,
            isbn=false,
            doi=false,
            backref=false,
            style=ieee,
            citestyle=numeric-comp,
            sorting=nyt,%none
            block=none]{biblatex}
\usepackage[font=small,labelfont=bf]{caption}
\usepackage{xspace}
\usepackage[bookmarks=true]{hyperref}
\hypersetup{
  colorlinks = true, %Colours links instead of ugly boxes
  urlcolor = Blue, %Colour for external hyperlinks
  linkcolor = Blue, %Colour of internal links
  citecolor = Blue, %Colour of citations
}

\renewcommand{\bibfont}{\small}

\newcommand{\del}[1]{\xspace}
\newcommand{\model}[0]{MBA\xspace}
\begin{document}
\title{\LARGE \bf Motion Before Action: \\ Diffusing Object Motion as Manipulation Condition}
\author{Yue Su$^{*1,2}$, Xinyu Zhan$^{*1}$, Hongjie Fang$^{1}$, Yong-Lu Li$^{1}$, Cewu Lu$^{1}$, and Lixin Yang$^{1 \dagger}$
\thanks{$^{*}$Equal Contribution.}
\thanks{$^{1}$Shanghai Jiao Tong University. $^{2}$Xidian University (this work was done while Y. Su is a research intern at Shanghai Jiao Tong University).}
\thanks{$^{\dagger}$Lixin Yang is the corresponding author. He is with the School of Artificial Intelligence, Shanghai Jiao Tong University, Shanghai, 200230, China}
\thanks{Author emails: s3702681@gmail.com, \{kelvin34501, galaxies, yonglu\_li, lucewu, siriusyang\}@sjtu.edu.cn.}
}

\markboth{}%
{How to Use the IEEEtran \LaTeX \ Templates}

\maketitle

\begin{abstract}
Inferring object motion representations from observations enhances the performance of robotic manipulation tasks.
This paper introduces a new paradigm for robot imitation learning that generates action sequences by reasoning about object motion from visual observations.
We propose \model, a novel module that employs two cascaded diffusion processes for object motion generation and robot action generation under object motion guidance.
\model first predicts the future pose sequence of the object based on observations, and then uses this sequence as a condition to guide robot action generation. 
Designed as a plug-and-play component, \model can be flexibly integrated into existing robotic manipulation policies with diffusion action heads. 
Extensive experiments in both simulated and real-world environments demonstrate that our approach substantially improves the performance of existing policies across a wide range of manipulation tasks. Project page: \url{https://selen-suyue.github.io/MBApage/} 
\end{abstract}

\section{Introduction}
Physiological research shows that humans process complex object motion information in their environment to support effective action execution~\cite{Human}. This motion analysis enables an understanding of object dynamics, which in turn guides human actions such as reaching, grasping, and maneuvering around obstacles.

In contrast to the biological approach, most existing robot policies~\cite{ACT, octo, DP, RISE, cage, DP3} are predominantly guided by observation, employing feature encoders and adopting generative approaches to predict actions. While effective, it often results in an overreliance on environmental cues, with the model focusing on memorizing observation features rather than reasoning about object motion patterns, as humans do. Consequently, when encountering extensive pose shifts in real-world objects or actions, many policies often struggle to generalize effectively~\cite{octo, openvla}, which can limit their practical performance.

To address these challenges and improve execution capabilities, we aim to equip the robot with human-like reasoning skills by \textit{inferring future object motion from observations}, and then \textit{predicting future actions under the object motion guidance}.
By achieving these objectives, the robot can derive intrinsic information (such as poses and motions) from the scene, allowing the policy to map observations to actions in a way that aligns more closely with human reasoning, \textit{i.e.}, reasoning about object motion rather than simply memorizing actions~\cite{jannerreasoning}.

\begin{figure}[!t]
    \centering
    \includegraphics[width=0.9\linewidth]{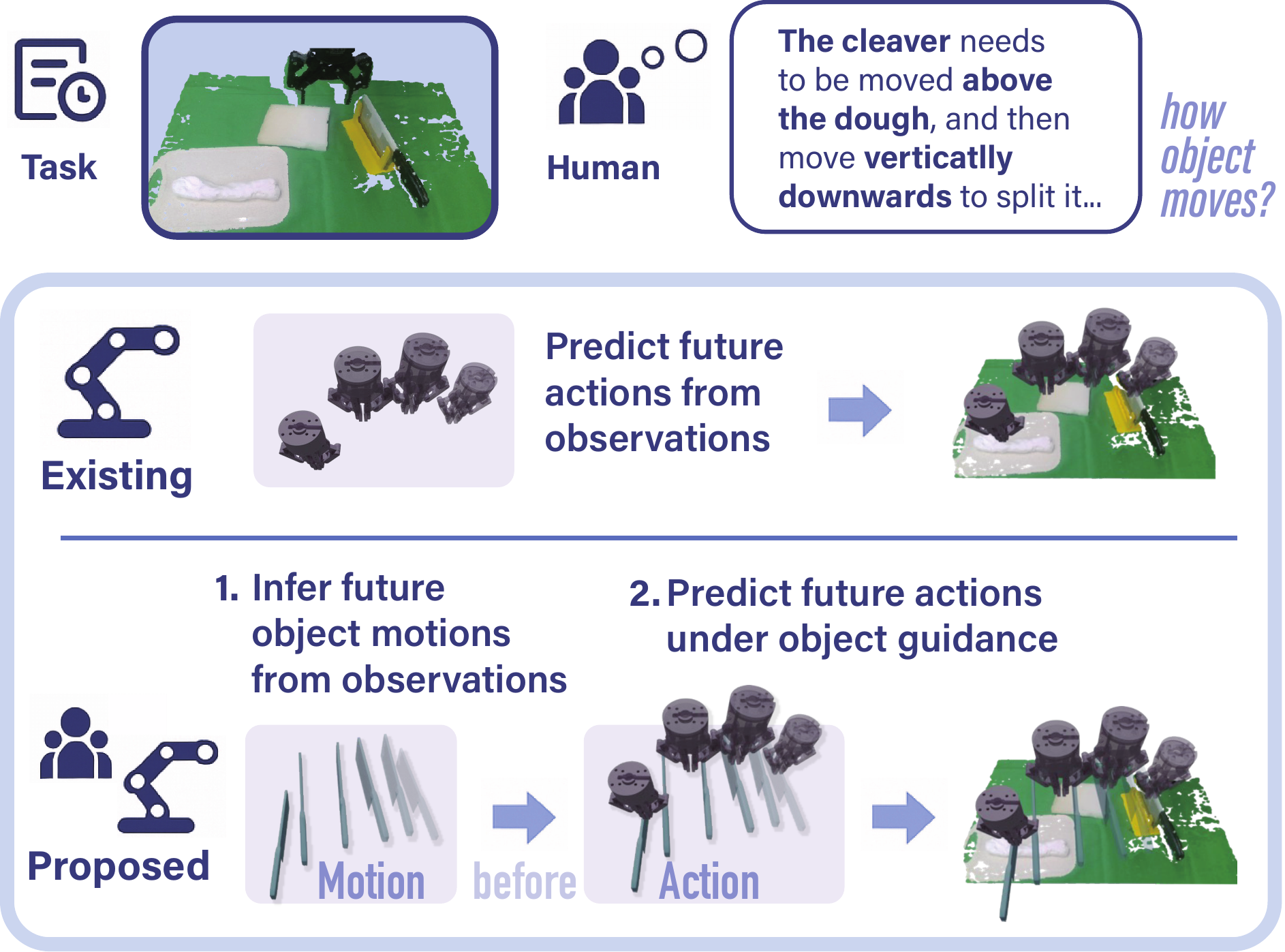}
    \caption[]{\red{\textbf{Understanding object motion before action leads to better manipulation}.} \red{Unlike existing methods that predict actions directly from observations, our approach first infers future object motions, enabling more accurate and goal-driven action prediction.}}
    \label{fig:teaser}
\end{figure}

In this work, we introduce \textbf{M}otion \textbf{B}efore \textbf{A}ction (referred to as \textbf{\model}), a novel module that equips existing robotic manipulation policies with such reasoning skills.
% use motion corresponding to the title and avoid repetition
We observe that representing the object motion using 6D pose aligns with the robot's end-effector pose in the same format, making the actions compatible with this representation. 
Consequently, they exhibit mathematical consistency.
During task execution, at each time step, the robot pose, object pose, and action are in proximity within the same space, with their kinematic relationships following a learnable pattern, demonstrating physical consistency~\cite{xu2021end}.
These consistencies suggest that the object poses and the robot poses and actions share similar distributions. 
The ability of probabilistic models, exemplified by diffusion models, to learn such similar distributions is transferable, as indicated in~\cite{DDPM,DDIM}.
Therefore, we propose that object poses can be generated by the diffusion process in the same way as actions.
These object poses can also effectively guide the action denoising process as conditions, just as robot poses do.

Building on this foundation, we design \model as a cascade of two diffusion modules. 
\model seamlessly integrates with any existing diffusion-based policy. 
The observation features encoded by the given policy serve as conditional inputs to the first diffusion module, which predicts future object pose sequences.
These predicted pose sequences, combined with the initial observations, are then fed as joint conditions to the diffusion action head, effectively guiding the policy's action generation.

We conduct comparative experiments with \model on three 2D and 3D robotic manipulation policies with diffusion action heads~\cite{DP, DP3, RISE}, demonstrating substantial performance improvements across various tasks. These tasks, comprising 57 tasks from 3 simulation benchmarks and 4 real-world tasks, involve articulated object manipulation, soft and rigid body manipulation, tool use, non-tool use, and diverse action patterns. Results show that \model consistently enhances the performance of such policies in both simulated and real-world environments.

In this paper, we make the following key contributions:
\begin{itemize}[leftmargin=10pt]
    \item We propose a novel imitation learning paradigm for robotic manipulation that allows robots to extract object pose sequences from observations and use them to aid in action prediction, thereby enhancing the robustness and kinematic consistency of the policy's observation-to-action mapping.
    \item We introduce \model, a flexible auxiliary module that can be easily integrated into existing policies with diffusion heads, rather than functioning as an independent policy. Extensive experiments demonstrate its ability to significantly improve the performance of these policies, highlighting its broad applicability and potential to enhance a wide range of robotic manipulation tasks.
\end{itemize}

\section{Related Work}
\subsection{Imitation Learning for Robotic Manipulation}

Imitation learning seeks to enable robots to acquire complex skills by learning from expert demonstrations. Recently, behavior cloning~\cite{bc} from demonstrations has demonstrated promising performance across various robotic manipulation tasks~\cite{BCZ, rt1, rt2, oxe, octo, openvla}. Most existing robotic manipulation policies are typically structured into three main components: an observation perception module for manipulation-relevant information extraction from observations, an optional policy backbone for processing the perception results, and an action head for generating actions based on the (processed) perception outputs.

For action generation, prior work has investigated both single- and multi-step action prediction methods. Single-step action prediction~\cite{rt1, rt2, oxe, openvla, zhang2024leveraging} is straightforward but often leads to inconsistencies between the consecutive actions. To address this limitation, several studies have applied the action chunking technique~\cite{ACT} to enable multi-step action sequence prediction. This includes the utilization of unimodal action heads, such as L1 and L2 action heads~\cite{ACT, roboagent, liu2024foam, kareer2024egomimic}, and multimodal action heads like the Gaussian mixture head~\cite{MandlekarWhat, bet, vqbet, cuiplay} and diffusion head~\cite{octo, DP, RISE, cage, DP3}. Among these, the diffusion head~\cite{DDPM} excels in capturing the diversity and complexity of action sequences, which is essential for executing real-world tasks.

Similar to the multimodal nature of robot action sequences in manipulation tasks, object motion sequences also exhibit diverse and complex characteristics, making diffusion heads well-suited for modeling such dynamics. Accordingly, we develop the \model module, designed for seamless integration into policies with diffusion action heads~\cite{DP, DP3, RISE}. This module enhances performance by decoupling the process into two stages: generating object motion via a motion diffusion head and conditionally generating robot actions through an action diffusion head.

\subsection{Object-Centric Manipulation Learning}\label{sec:related-works-object-centric}

Object-centric manipulation learning can be categorized into two main lines of approaches: one emphasizing scene understanding through structured, object-centric representations~\cite{groot, viola, orion, tyree20226, migimatsu2020object, devin2018deep}, and the other focusing on action-oriented learning by identifying how objects can be manipulated to achieve specific goals~\cite{bahl2023affordances, sriramahrp, XuFlowCrossDomain, BharadhwajTrack2act, YuanGeneralFlow}.

The action-oriented approach actively engages with objects to achieve desired outcomes. Affordance learning is a central method here, aiming to identify possible actions based on an object’s physical properties and context~\cite{bahl2023affordances, SAM2_policy, BorjaAffordance, nasiriany2024rt, yuan2024robopoint, sriramahrp}. This approach, though effective in structured scenarios, often relies on predefined affordances, limiting its adaptability to dynamic environments and tasks that require fine-grained coordination of actions or adaptive responses.

To overcome these limitations, inspired by the success of flow-based methods~\cite{teed2020raft, goyal2022ifor}, recent work has utilized object flow as a general affordance for guiding manipulation policies~\cite{ATM, VecerikRobotap, XuFlowCrossDomain, BharadhwajTrack2act, YuanGeneralFlow, EisnerFlowbot3d, ZhangFlowbotPlusPlus, SeitaToolflownet}. These policies map vision to motion by anchoring visual features and constructing object flow, which is then translated into robot actions using either heuristic transformations~\cite{YuanGeneralFlow, BharadhwajTrack2act, ZhangFlowbotPlusPlus, SeitaToolflownet, EisnerFlowbot3d} or learned models~\cite{ATM, XuFlowCrossDomain}. Many methods incorporate object pose as an intermediate representation~\cite{YuanGeneralFlow, BharadhwajTrack2act, SeitaToolflownet} in such flow-to-action translations. However, using flow as an indirect motion representation can introduce ambiguity due to the vision-motion gap~\cite{ATM, XuFlowCrossDomain, BharadhwajTrack2act}, complicating flow-to-action learning. Moreover, defining such heuristic transformations is challenging, often hindering generalization in dynamic or novel environments.

Our proposed \model module employs object pose as a direct motion representation, which reduces training demands and enhances interpretability by establishing a clear link between visual inputs and motion outcomes. While a contemporaneous work~\cite{SPOT} similarly leverages object pose for motion representation, it infers robot actions heuristically from object trajectories. We argue that generating actions from object motion guidance is essential, particularly in tasks involving deformable objects or dynamic conditions where heuristic methods may fall short.
            
\section{Method}

We aim to endow the policy with the capability to concurrently reason about object pose $\mathbf{M}$ and robot action $\mathbf{A}$ from observations $\mathbf{O}$.
Specifically, we model the joint conditional distribution $p(\mathbf{M},\mathbf{A}|\mathbf{O})$ in \model.
This distribution can be decoupled as:
$$
p(\mathbf{M},\mathbf{A}|\mathbf{O}) = p(\mathbf{M}|\mathbf{O})p(\mathbf{A}|\mathbf{M},\mathbf{O})
$$
where $p(\mathbf{M}|\mathbf{O})$ serves to sample the most likely object motions based on current observations, and $p(\mathbf{A}|\mathbf{M},\mathbf{O})$ serves to sample the most likely actions, guided by the sampled motions and current observations.

\begin{figure*}[!ht]
    \centering
    \includegraphics[width=0.9\textwidth]{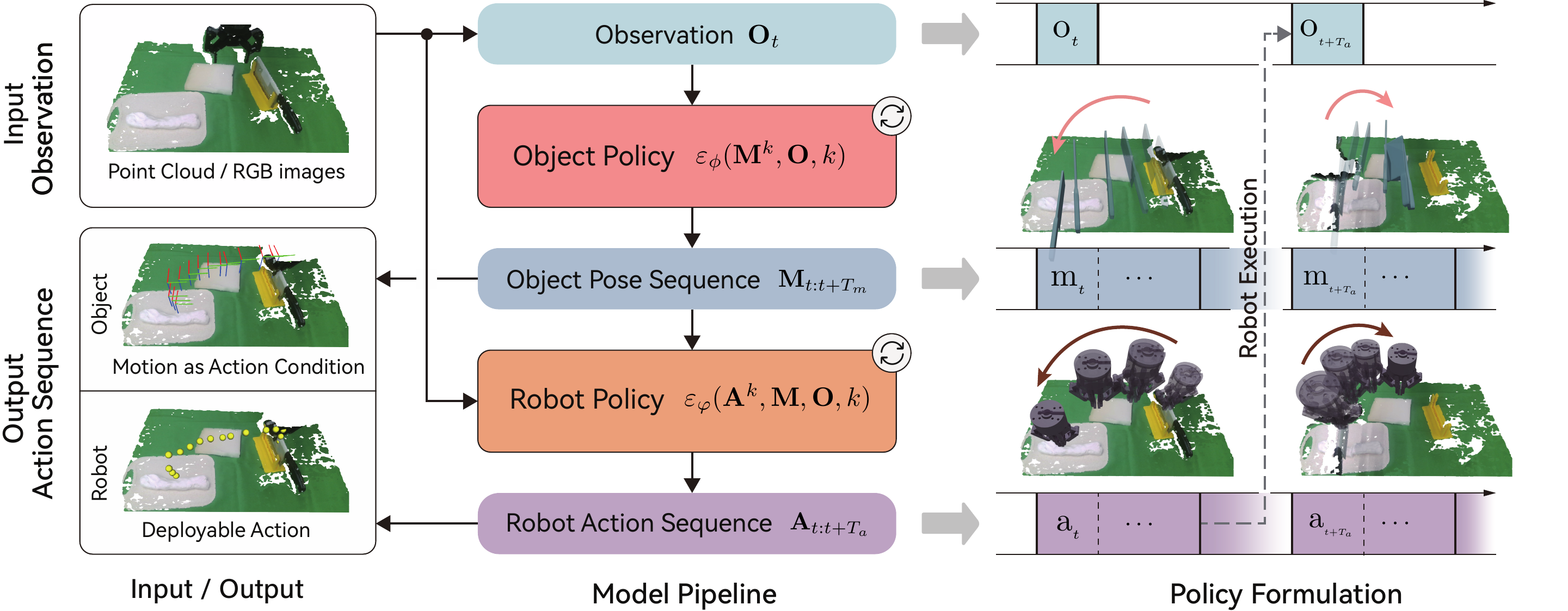}
    \caption{\textbf{Overview of \model pipeline}. \model takes the current observation as input, which could be in the form of 3D point clouds or RGB images from different viewpoints. 
    Object pose sequences are sampled as actions with denoising diffusion from the object policy to be part of the framework's output. 
    Conditioning on the observations and object pose actions, \model samples deployable robot actions with denoising diffusion from the robot policy. These actions are executed within the workspace to update the environment state and the observations.}
    \label{fig:pipeline}
\end{figure*}

\model has an object motion generation module $p(\mathbf{M}|\mathbf{O})$ that is readily compatible with any existing policy with diffusion action heads, enhancing versatility and efficiency across various robotic manipulation contexts. Given an observation, the perception module of the original policy first processes it into observation features. These observation features are then passed to \model, where they serve as a condition for the object pose sequence denoising process. Subsequently, the diffusion action head of the original policy takes both the observation features and the denoised object pose sequence as inputs and outputs the corresponding robot actions. During training, both ground-truth object pose sequences and action sequences are used as supervision for the predicted object pose sequences from \model and action sequences from the action head. At inference time, the policy with \model operates in an end-to-end manner, without requiring ground-truth object poses. The full execution framework of \model is illustrated in Fig.~\ref{fig:pipeline}. 

\subsection{Object Motion Generation}
As discussed in \S\ref{sec:related-works-object-centric}, we opt for object pose as a direct representation to describe object motion. The object pose $\mathrm{m}_t$ is represented as a 9D vector, combining 3D translation and 6D rotation~\cite{rot6d}, to align with robot action representations~\cite{DP, DP3, RISE}.
At each observation step $t$, we use the observation features $O_t$ as the condition to the diffusion process~\cite{DDPM}, which denoises the object pose sequence for the next $T_m$ steps $\mathbf{M}_{t:t+T_m} = (\mathrm{m}_t,\mathrm{m}_{t+1},\mathrm{m}_{t+2} ... \mathrm{m}_{t+T_m-1})$.

Specifically, the diffusion-based object pose generation model begins by sampling an initial noise from a Gaussian distribution $\mathbf{M}^k \sim \mathcal{N}(0, I)$.
At each diffusion step $k$, the denoising network $\varepsilon_\theta$ progressively removes this noise, conditioned on the observation features $O_t$. This process iterates to recover a noise-free, clean object pose sequence $\mathbf{M}^0$:
\begin{equation}
    \mathbf{M}^{k-1} = \alpha_k \left( \mathbf{M}^k - \gamma_k\varepsilon_\phi(\mathbf{M}^k, O, k) \right) + \sigma_k \mathcal{N}(0, I),
\end{equation}
where $\{\alpha_k,\gamma_k,\sigma_k\}$ is the noise schedule.

To train the noise prediction network $\varepsilon_\theta$, we randomly pick a sequence of real object trajectory $\mathbf{M}^0$ and apply noise at a random diffusion step $k$ through a forward diffusion process.
The model is then tasked to predict the added noise $\varepsilon^k$ at the corresponding diffusion step.
We employ mean squared error (MSE) loss as the objective function to supervise the prediction of object pose:
\begin{equation}
    \mathcal{L} = \text{MSE}\left(\epsilon^k, \varepsilon_\phi(\mathbf{M}^0 + \epsilon^k, {O}, k)\right),
\end{equation}
where the ground-truth object pose $\mathbf{M}^0$ is obtained via a motion capture (MoCap) system during demonstration collection, and $\epsilon^k$ is the added noises for the diffusion step $k$.

\subsection{Robot Action Generation under Object Motion Guidance}
Following~\cite{DP,DP3,RISE}, the robot action $\mathrm{a}_t$ is defined by the end-effector pose and gripper width, using a 10D representation that concatenates 3D translation, 6D rotation~\cite{rot6d}, and 1D gripper width together.
%The action the robot is set to perform at the current moment $\mathrm{a}_t$ is represented as:
%\begin{equation}
%    \mathrm{a}_t= (x^{t+1},y^{t+1},z^{t+1},r^{t+1}_1\ ...\ r^{t+1}_6,g^{t+1})
%\end{equation}
Similarly, we use the diffusion process to generate the action sequence at the next $T_a$ steps: $\mathbf{A}_{t:t+T_a} = (\mathrm{a}_t,\mathrm{a}_{t+1},\mathrm{a}_{t+2} ... \mathrm{a}_{t+T_a-1})$.

In this process, we generate actions $\mathbf{A}^0$ using the same diffusion method as for object poses. 
Notably, the action noise prediction network receives not only the same observation features $O$ as before, 
but also the object pose features $M$ obtained in the previous stage. These features are concatenated to predict the noise jointly:
\begin{equation}
    \mathbf{A}^{k-1} = \alpha_k \left( \mathbf{A}^k - \gamma_k\varepsilon_\phi(\mathbf{A}^k, M, O, k) \right) + \sigma_k \mathcal{N}(0, I)
\end{equation}

Given the physical and mathematical similarities between object poses and the robot's end-effector poses, we use a similar encoding network for their feature representation. 
The object pose is unfolded along the \( T_m \) axis into \( T_m \times 9 \) and then passed through MLP layers with dimensions \((T_m \times 9, 32, 32)\), 
ultimately resulting in the object pose feature vector \( M \).

From every iteration, we collect an action sequence \( \mathbf{A}^0 \) from expert demonstrations and add noise over \( k \) steps. 
The noise prediction network is then tasked with predicting the noise based on the feature information, 
with supervision provided through MSE loss:
\begin{equation}
    \mathcal{L} = \text{MSE}\left(\epsilon^k, \varepsilon_\varphi(\mathbf{A}^0 +  \epsilon^k, {M}, {O}, k)\right)
\end{equation}

\begin{algorithm}[!htp]
    \small
    \caption{\model Execution Process}
    \label{alg:load_execute}
    \begin{algorithmic}
        % \setstretch{0.83} 
        \REQUIRE Observation \( \mathbf{O}_t \), Predetermined steps \( T_m, T_a' \leq T_a \), Encoders \( \mathbf{F}_{\mathrm{O}}, \mathbf{F}_{\mathrm{M}} \)
        \ENSURE Task completion via iterative action execution
        \WHILE{task is not complete}
            \STATE Encode the observation: $O_t \gets \mathbf{F}_{\mathrm{O}}(\mathbf{O}_t)$
            \STATE Predict object pose sequence \( \{ \mathrm{m}_t, \mathrm{m}_{t+1}, \dots, \mathrm{m}_{t+T_m} \} \):
            \FOR{$ k = K $ downto $1$}
                \STATE $
                \mathbf{M}^{k-1} = \alpha_k \left( O^k - \gamma_k \epsilon_\phi(\mathbf{M}^k, O_t, k) \right) + \sigma_k \mathcal{N}(0, I)
                $
            \ENDFOR
            \STATE $M_t \gets \mathbf{F}_{\mathrm{M}} \left( \mathbf{M}^0 = \{ \mathrm{m}_t, \mathrm{m}_{t+1}, \dots, \mathrm{m}_{t+T_m} \} \right)$
            \STATE Predict action sequence \( \{ \mathrm{a}_t, \mathrm{a}_{t+1}, \dots, \mathrm{a}_{t+T_a} \} \) with extracted feature $M_t$ and $O_t$:
            \FOR{$ k = K $ downto $1$}
                \STATE $\mathbf{A}^{k-1} = \alpha_k \left( A^k - \gamma_k \epsilon_\varphi(\mathbf{A}^k, M_t, O_t, k) \right) + \sigma_k \mathcal{N}(0, I)$
            \ENDFOR
            \STATE Execute \( T_a' \) steps, \( T_a' \leq T_a \)
            \FOR{$i = 1$ to \( T_a' \)}
                \STATE Execute \( \mathrm{a}_{t+i} \):
                $
                    \mathrm{a}_{t+i} \gets 
                    \begin{cases} 
                        \mathrm{a}_{t+i} & \text{If task condition holds} \\
                        \text{Stop} & \text{Otherwise}
                    \end{cases}
                $
            \ENDFOR
            \STATE Update observations \( \mathbf{O}_t \)
            \STATE $t \gets t + T_a'$
        \ENDWHILE
    \end{algorithmic}
\end{algorithm}

\subsection{Execution}
During execution, the policy with \model estimates the object pose for \( T_m \) steps based on the observation inputs, which include environmental information and the robot state. 
This estimated pose sequence serves as a direct condition for generating an action sequence of \( T_a \) steps, without reliance on the MoCap system. The robot then executes up to \( T_a \) action steps before the next observation and execution cycle, 
with the exact number of steps depending on task type and environmental conditions. To adhere to standard kinematic principles, we ensure that \( T_m \geq T_a \) in all experiments. The specific process is detailed in Alg.~\ref{alg:load_execute}.

\section{Simulation Experiments}

In the simulation experiments, we aim to address the following research questions: \textbf{(Q1)} Can \model effectively improve the performance of robotic manipulation policies by leveraging predicted object motion as a condition for generating robot actions? \textbf{(Q2)} Can the human-like reasoning process of \model make policy learning more efficient?

\begin{table*}[!t]
    \centering
    \renewcommand\arraystretch{1.6}
    \setlength\tabcolsep{2pt} 
    \footnotesize
    \begin{tabularx}{0.84\textwidth}{c>{\centering \arraybackslash}X>{\centering\arraybackslash}X>{\centering\arraybackslash}X>{\centering\arraybackslash}X>{\centering\arraybackslash}X>{\centering\arraybackslash}X>{\centering\arraybackslash}X}
    \hline
     \textbf{Method}& \textit{\textbf{Adroit}} (3) & \textit{\textbf{DexArt}} (4) & \textit{\textbf{MetaWorld} \newline Easy} (28) & \textit{\textbf{MetaWorld} Medium} (11) & \textit{\textbf{MetaWorld} \newline Hard} (6) & \textit{\textbf{MetaWorld} VeryHard} (5) & \textbf{Average} (57) \\
    \hline
   
    DP3~\cite{DP3} & $68.3\pm 3.3$ & \cellcolor[HTML]{acbed2} $53.5\pm 2.5$ & $90.9\pm1.4$ & $61.6\pm6.5$ & $29.7\pm 2.8$ & $49.0\pm 6.8$ & $71.3\pm3.2$  \\
    DP3 \textit{w.} \model & \cellcolor[HTML]{acbed2}$79.7\pm 0.7$ & $52.3\pm 2.8$ & \cellcolor[HTML]{acbed2} $92.5\pm1.1$ & \cellcolor[HTML]{acbed2} $66.4\pm6.1$ & \cellcolor[HTML]{acbed2} $36.2\pm1.2$ & \cellcolor[HTML]{acbed2} $86.8\pm 1.6$ & \cellcolor[HTML]{acbed2} $77.5\pm2.2$  \\
    \hline
  
    DP~\cite{DP} & $31.7\pm 3.0$ & $22.8\pm 3.8$ & $83.6\pm3.6$ & $31.1\pm5.3$ & $9.0 \pm 1.0$ & $26.6 \pm 3.2$& $53.6\pm3.6$   \\
    DP \textit{w.} \model & \cellcolor[HTML]{acbed2} $64.0\pm 3.0$ & \cellcolor[HTML]{acbed2} $29.0\pm 2.5$ & \cellcolor[HTML]{acbed2} $84.9\pm1.2$ & \cellcolor[HTML]{acbed2} $57.8\pm4.4$ & \cellcolor[HTML]{acbed2} $23.8\pm 1.3$ & \cellcolor[HTML]{acbed2} $79.8 \pm 1.2$ & \cellcolor[HTML]{acbed2} $67.8\pm2.0$  \\
    \hline
    \end{tabularx}
    \caption{\textbf{Performance comparison of policy models with and without MBA integration across 57 simulation tasks.} Average success rates ($\pm$ standard deviation) are reported under three random seeds.}
    \label{sim_result}
\end{table*}

\subsection{Setup}
\textbf{Benchmarks.} We evaluate our policy in three simulation benchmarks, encompassing a total of 57 tasks:
\begin{itemize}[leftmargin=10pt]
\item \textit{\textbf{Adroit}}~\cite{Adroit} utilizes a multi-fingered Shadow robot in the MuJoCo~\cite{mujoco} environment to perform highly dexterous manipulation 
across a variety of tasks. These operations involve both articulated objects and rigid bodies.
\item \textit{\textbf{DexArt}}~\cite{dexart} uses the Allegro robot in the SAPIEN~\cite{sapien} environment to perform high-precision dexterous manipulation. 
It primarily focuses on tasks involving articulated object manipulation.
\item \textit{\textbf{MetaWorld}}~\cite{metaworld} primarily operates in the MuJoCo environment, using a gripper to perform manipulation tasks involving both articulated and rigid objects. It covers a wide range of skills required for everyday scenarios, categorizing these tasks into difficulty levels: easy, medium, hard, and very hard.
\end{itemize}

\textbf{Baselines.} The focus of this work is to demonstrate that the introduction of \model as a module can universally enhance the performance of existing policies with diffusion heads. Therefore, in our simulation experiments, we select representative Diffusion Policy (\textbf{DP})~\cite{DP}, 3D Diffusion Policy (\textbf{DP3})~\cite{DP3}
%and \red{\textbf{RISE}~\cite{RISE}} 
as our 2D and 3D baselines. 
%\red{For RISE, since its model architecture is specifically intended for real-world scenarios rather than simulation environments~\cite{RISE}, we reimplement RISE in simulation.} 
We then integrate the \model module into these baselines and compare the performance. 
To ensure fairness in the experiments, we ensure that \model and the corresponding baseline methods use the same expert demonstrations for training, with an equal amount of training steps. During the execution phase, both methods undergo the same number of observation and inference steps.

\textbf{Demonstrations.} In terms of expert demonstration generation, we employ scripted policies for \textbf{\textit{MetaWorld}}, the VRL3~\cite{vrl3} agent for \textbf{\textit{Adroit}}, and the PPO~\cite{ppo} agent for \textbf{\textit{DexArt}}. The average demonstration success rates for these agents are 98.7\%, 72.8\%, and nearly 100\%, respectively. For \textbf{\textit{Adroit}} and \textbf{\textit{MetaWorld}}, we use 10 demonstrations for training, while 100 demonstrations are used for \textbf{\textit{DexArt}}.

\textbf{Protocols.} Following \cite{DP3}, We conduct 3 runs for each experiment, using seed numbers 0, 1, and 2. \red{For each seed, we evaluate 20 episodes every 200 training epochs as a test node, compute the average success rate over 20 episodes for each test node, and then average the top 5 performing test nodes. 
This accounts for the inherent instability of imitation learning policies during training in simulated environments, where different policies converge at varying training stages. We then report the mean and standard deviation of success rates across the 3 seeds.}

\begin{table*}[!t]
    \centering
    \renewcommand\arraystretch{1.4}
    \setlength\tabcolsep{3pt}
    \setlength{\aboverulesep}{0pt}
    \setlength{\belowrulesep}{0pt}
    \footnotesize
    \begin{tabular}{cccccccccccc}
        \hline
        \multirow{2}{*}{\textbf{Method}}  & \multicolumn{2}{c}{\textit{\textbf{Adroit}}} & \multicolumn{2}{c}{\textit{\textbf{DexArt}}} & \multicolumn{7}{c}{\textit{\textbf{MetaWorld}}} \\
        \cmidrule(r){2-3}\cmidrule{4-5}\cmidrule(l){6-12}
        & \textit{Door} & \textit{Pen} & \textit{Laptop} & \textit{Toilet} & \textit{Bin Picking} & \textit{Box Close} & \textit{Hammer} & \textit{Peg Insert Side} & \textit{Disassemble} & \textit{Shelfplace} & \textit{Reach} \\
        \hline
        DP3~\cite{DP3} & $62\pm4$ & $43\pm6$ & \cellcolor[HTML]{acbed2}$81\pm2$ & \cellcolor[HTML]{acbed2}$71\pm3$ & $34\pm30$ & $42\pm3$ & $76\pm4$ & $69\pm7$ & $69\pm4$ & $17\pm10$ & $24\pm1$ \\
        DP3 \textit{w.} \model & \cellcolor[HTML]{acbed2}$74\pm1$ & \cellcolor[HTML]{acbed2}$65\pm1$ & $78\pm6$ & $70\pm2$ & \cellcolor[HTML]{acbed2}$54\pm23$ & \cellcolor[HTML]{acbed2}$56\pm2$ &\cellcolor[HTML]{acbed2}$98\pm2$ & \cellcolor[HTML]{acbed2}$75\pm5$ & \cellcolor[HTML]{acbed2}$98\pm1$ & \cellcolor[HTML]{acbed2}$73\pm1$ & \cellcolor[HTML]{acbed2}$32\pm5$ \\
        %\hline
        %\red{RISE$^*$~\cite{RISE}} &  \red{$44\pm2$} & \red{$21\pm1$} &\red{$46\pm2$} &\red{$58\pm5$} & \red{$24\pm5$} & \red{$45\pm3$} & \red{$85\pm3$} & \red{$33\pm4$} & \red{$64\pm4$} & \red{$13\pm3$} & \red{$18\pm2$} \\
        %\red{RISE$^*$ \textit{w.} \model} & \red{$66\pm2$} & \red{$42\pm1$} &\red{$80\pm1$} &\red{$67\pm5$} & \red{$51\pm19$} & \red{$54\pm8$} & \cellcolor[HTML]{acbed2}\red{$100\pm0$} & \red{$71\pm2$} & \cellcolor[HTML]{acbed2}\red{$100\pm0$} & \red{$36\pm3$} & \red{$20\pm0$}  \\
        \hline
        DP~\cite{DP} & $37\pm2$ & $13\pm2$ & $31\pm4$ & $26\pm8$ & $15\pm4$ & $30\pm5$ & $15\pm6$ & $34\pm7$ & $43\pm7$ & $11\pm3$ & $18\pm2$\\
        DP \textit{w.} \model & \cellcolor[HTML]{acbed2}$45\pm1$ & \cellcolor[HTML]{acbed2}$53\pm3$ & \cellcolor[HTML]{acbed2}$42\pm2$ & \cellcolor[HTML]{acbed2}$45\pm2$ & \cellcolor[HTML]{acbed2}$45\pm8$ & \cellcolor[HTML]{acbed2}$36\pm7$ & \cellcolor[HTML]{acbed2}$89\pm5$ & \cellcolor[HTML]{acbed2}$53\pm0$ & \cellcolor[HTML]{acbed2}$71\pm23$ & \cellcolor[HTML]{acbed2}$64\pm2$ & \cellcolor[HTML]{acbed2}$26\pm3$ \\
        \hline
    \end{tabular}
    \caption{\textbf{Task-level success rates across 11 simulation tasks in 3 environments} comparing MBA-augmented and baseline policies.}
    \label{simtask}
\end{table*}
\vspace{1em}
\begin{figure*}[!t]
    \centering
    \subfloat{\includegraphics[width=0.16\textwidth]{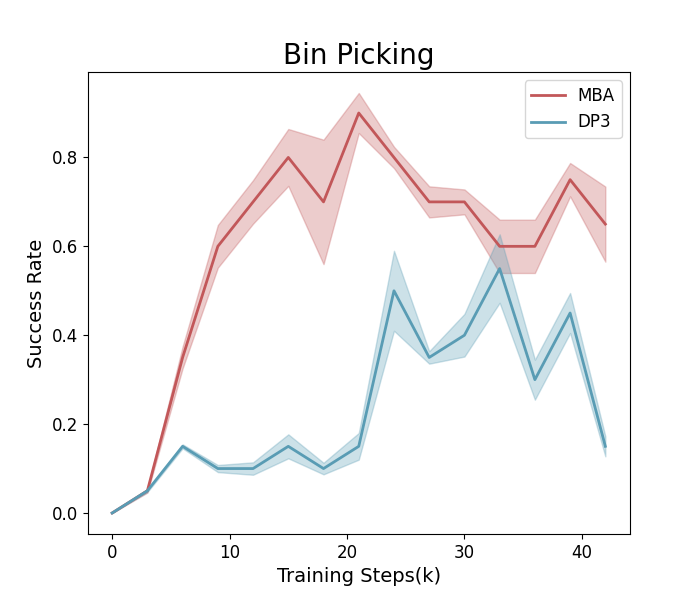}}%
    \hfil
    \subfloat{\includegraphics[width=0.16\textwidth]{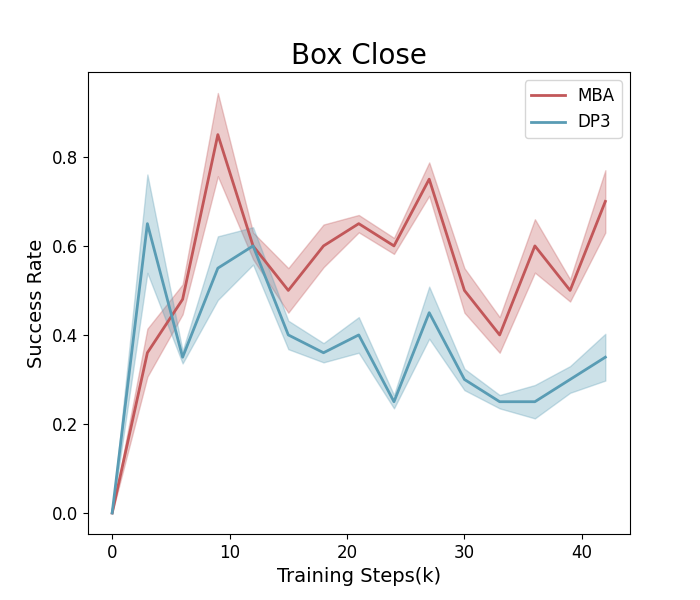}}%
    \hfil
    \subfloat{\includegraphics[width=0.16\textwidth]{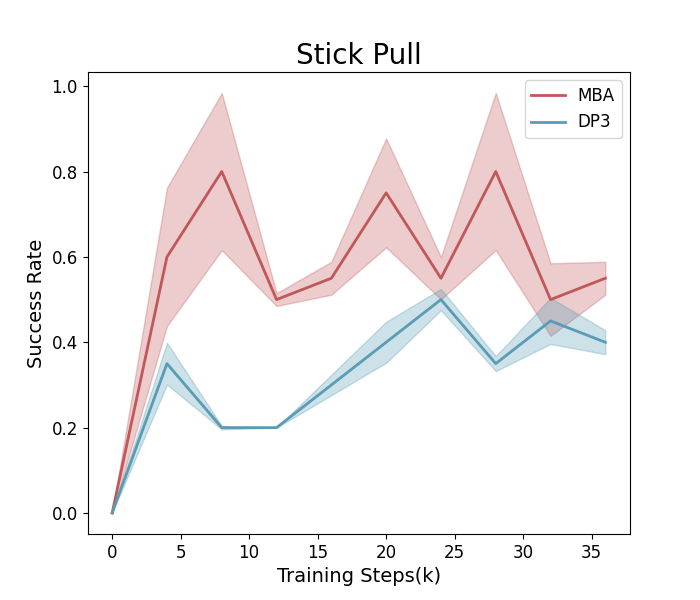}}%
    \hfil
    \subfloat{\includegraphics[width=0.16\textwidth]{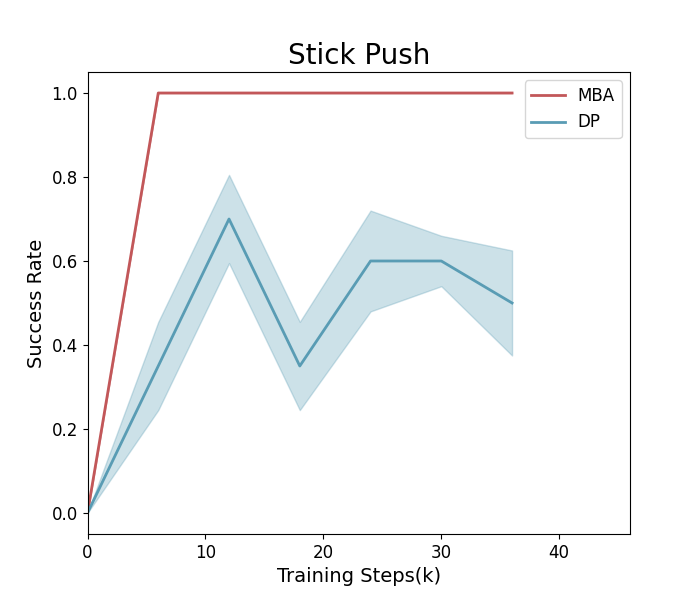}}%
    \hfil
    \subfloat{\includegraphics[width=0.16\textwidth]{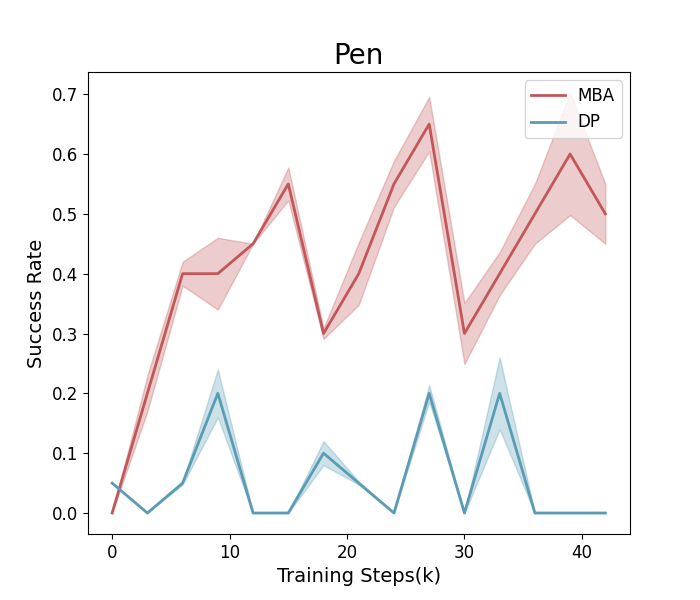}}%
    \hfil
    \subfloat{\includegraphics[width=0.16\textwidth]{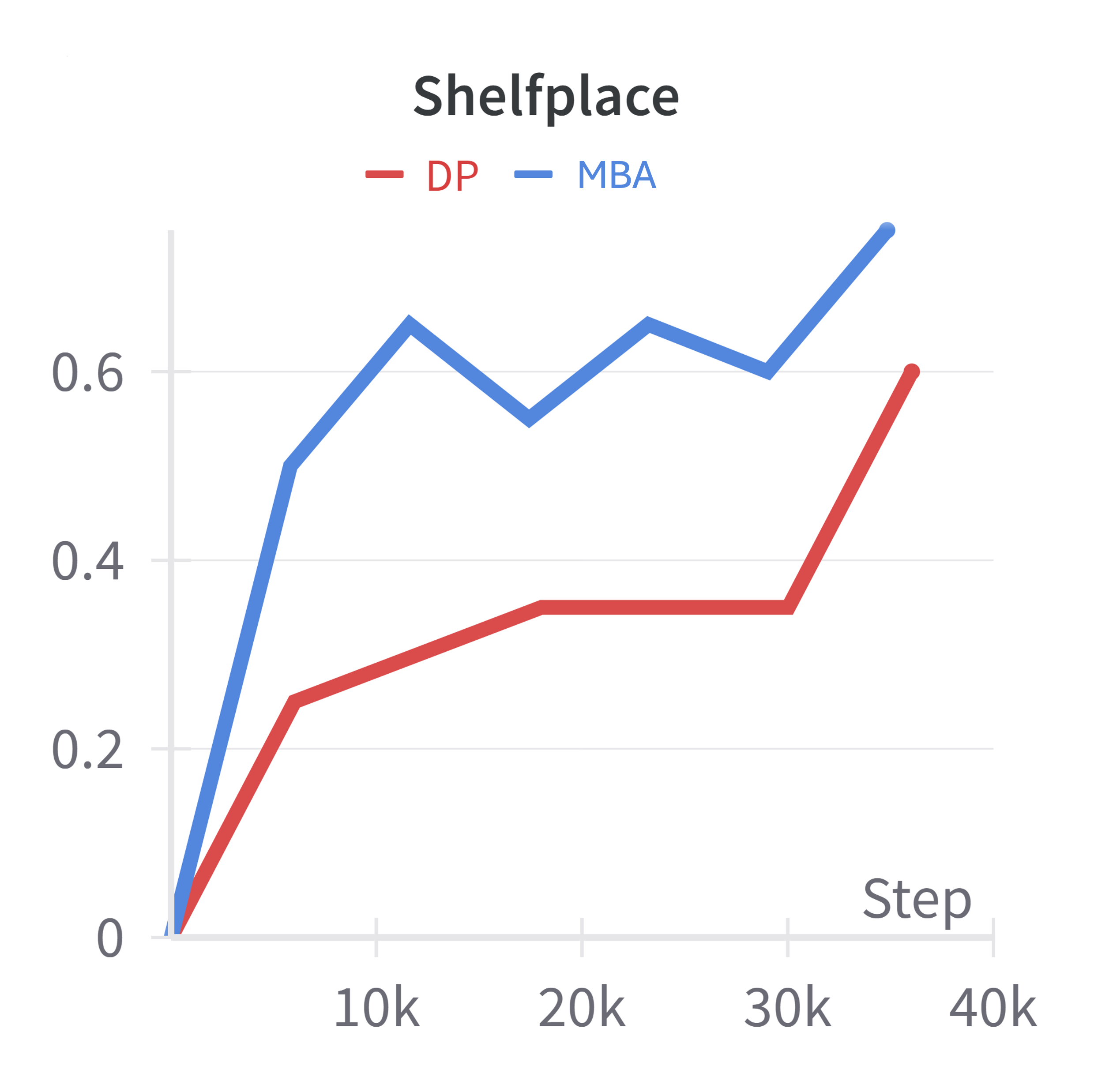}}%
    \caption{\textbf{Average learning curves (success rate - training steps)} over three runs  comparing MBA-augmented and baseline policies.}
    \label{efficient chart}
\end{figure*}

\subsection{Results}
\textbf{Integrating \model results in more stable and superior performance (Q1).} We report the average success rates and standard deviations (across three random seeds) for all simulation tasks in the Adroit, Dexart, and MetaWorld environments, across four difficulty levels, in Table.~\ref{sim_result}.
On average, \model outperforms the baseline in the majority of benchmarks, achieving a 14.2\% increase in the average success rate over DP and a 6.2\% increase over DP3.
Additionally, we observe a general reduction in the average standard deviation of task execution across all benchmarks, further confirming the robustness of \model.
The results also demonstrate that \model significantly improves success rates in tasks with higher difficulty levels.
Detailed examples are provided in Table.~\ref{simtask}, where we report the average success rate and standard deviation for particular tasks.
Notably, in tasks that require precise contact and fine manipulation within narrow action spaces, \model can enhance the execution capability and robustness of the robotic manipulation policies.
%\red{\model also achieves strong performance when built on the RISE's visual backbone.}
We attribute this success primarily to two factors.
First, when objects are stationary, \model 's accurate prediction (can also be regarded as a pose estimation in this situation) of object pose enables the robot to localize and grip the objects effectively.
Second, when objects are in motion, the strong correlation between the predicted object pose sequences and the robot's action sequences provides robust guidance and calibration for action generation.

\textbf{\model accelerates the policy learning process, leading to more efficient learning (Q2).} As shown in Fig.~\ref{efficient chart}, we observe that policies with \model exhibit higher learning efficiency during training compared to their vanilla counterparts. These policies typically reach the peak task test success rate at earlier training steps and maintain a more stable success rate at a higher level. This improvement is primarily attributed to the object pose information, which offers a more learnable and easily encoded feature representation.

\section{Real-World Experiments}
In the real-world experiments, we aim to evaluate the effectiveness of the proposed \model module in enhancing the performance of real-world robotic manipulation policies across a variety of manipulation tasks.

\subsection{Setup}
\begin{figure}[!t]
   \centering
   \includegraphics[width=0.49\textwidth,height=3.5cm]{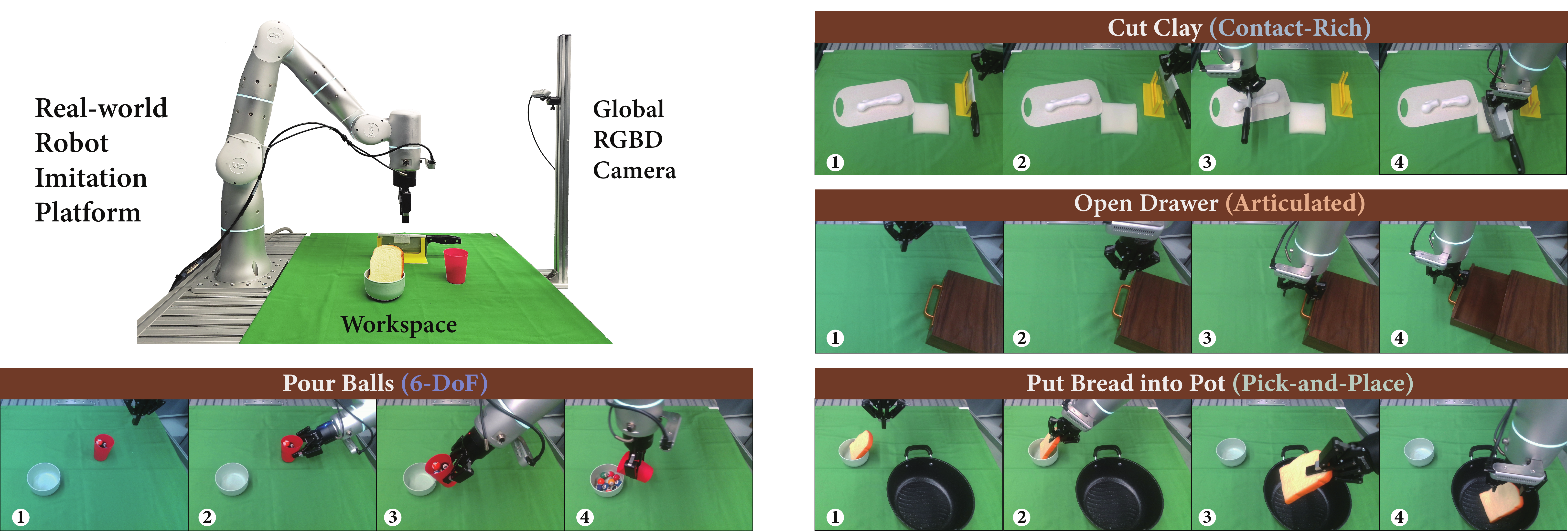}
    \caption{\textbf{Real-world deployment platform and execution process of four manipulation tasks.}}
    \label{load-qualitative}
\end{figure}

\textbf{Platform.} We employ a Flexiv Rizon robotic arm with a Robotiq 2F-85 gripper to manipulate objects. A global Intel RealSense D415 RGB-D camera is positioned in front of the robot for 3D perception, capturing single-view workspace point clouds. The robot workspace is defined as a 40 cm \(\times\) 60 cm rectangular area in front of the robot. The overview of the real-world robot platform is shown in Fig.~\ref{load-qualitative}.
All devices are connected to a workstation with an Intel i9-10980XE CPU and an NVIDIA 2080 Ti GPU for data collection and evaluation.
Since the training process of \model requires 6D object motion data for supervision, we set up a Motion Capture (MoCap) system comprising five OptiTrack Prime 13W infrared cameras.
\del{This system captures 6D object poses by tracking reflective hemisphere markers attached to each object.}
% Before the training phase, we inpaint those MoCap markers within RGB-D observations to maintain visual consistency between the training and deployment.
%\red{While deployment results indicate that omitting this step has little impact on \model's performance, we still recommend this practice to maintain consistency and minimize the risk of the model memorizing marker locations.}
Notably, \textit{no motion capture system or markers are required during deployment}.

\textbf{Tasks.} We select 4 tasks for experimentation: \textbf{\textit{Cut Clay}} (a tool-use contact-rich task), \textbf{\textit{Put Bread into Pot}} (soft body manipulation), \textbf{\textit{Open Drawer}} (articulated object manipulation),  and \textbf{\textit{Pour Balls}} (6-DoF task) as illustrated in Fig.~\ref{load-qualitative}.\

\textbf{Demonstrations.} We collect 50 expert demonstrations through end-effector teleoperation with haptic devices for each task, following the same setup and procedures in~\cite{rh20t, RISE, cage}.

\textbf{Baseline.} \red{We retain DP and DP3 from the previous section as baselines,} and additionally include  RISE~\cite{RISE}, a SOTA real-world policy model that operates solely on 3D scene input. RISE generates actions through a diffusion head, conditioned on 3D features extracted via sparse convolution networks~\cite{spatioenc} and transformers. We integrate the \model module into these three policies, placing it before the action head to predict the future object pose sequence as the action condition.

\textbf{Protocols.} In the real-world evaluation, we conduct 20 trials per method for each task unless stated otherwise.
All methods are compared under nearly identical randomized initial scene configurations for each trial.

\subsection{Cut Clay}
% \begin{table}[!h]
% \centering
% \renewcommand\arraystretch{1.2}
% \setlength\tabcolsep{3pt}
% \small
% \begin{tabular}{c c c c c c}
% \toprule
% \multirow{2}{*}{\textbf{Methods}} & \multicolumn{5}{c}{\textbf{Metrics}} \\
% \cmidrule{2-6}
% \ & {\footnotesize \textit{Pick} (\%)} & {\footnotesize \textit{Cut} (\%)} & {\footnotesize \textit{Seperate} (\%)} & {\footnotesize \textit{Place}(\%)} & {\footnotesize \textit{Attempt}$\downarrow$} \\
% \midrule
% RISE & $95$ & $70$ & $30$ & $50$ & \cellcolor[HTML]{acbed2}$11$\\
% RISE \textit{w.} \model &\cellcolor[HTML]{acbed2}$100$ &\cellcolor[HTML]{acbed2}$90$ &\cellcolor[HTML]{acbed2}$55$  &\cellcolor[HTML]{acbed2}$65$ & $12$ \\
% \midrule
% DP3 & \cellcolor[HTML]{acbed2}$95$ & $35$ & $10$ & $15$ & $53$\\
% DP3 \textit{w.} \model &$80$ &\cellcolor[HTML]{acbed2}$60$ &\cellcolor[HTML]{acbed2}$20$  &\cellcolor[HTML]{acbed2}$25$ & \cellcolor[HTML]{acbed2}$25$ \\
% \bottomrule
% \end{tabular}
% \caption{Experimental results of the \textbf{\textit{Cut Clay}} task.}
% \label{realtable_cut}
% \end{table}

% \vspace{1em}
The \textbf{\textit{Cut Clay}} task consists of three stages: grasping the knife, slicing the clay until separation, and placing the knife on a foam mat. This task involves multiple stages, testing the ability of \model to guide policy execution through accurate prediction of future object pose sequences.

In this experiment, we randomize the position and orientation of the cutting board, knife holder, and foam mat for each test.
Special attention is given to varying the relative positions of these three objects to assess policy robustness thoroughly.
Additionally, the shape of the clay is changed each time to meet the generalization requirements for real-world applications. We define five evaluation metrics for this task:

\begin{itemize}
    \item Success rate of picking the knife.
    \item Success rate of completing the cutting motion.
    \item Success rate of slicing the clay until separation.
    \item Success rate of placing the knife on the mat.
    \item Total redundant knife-grasp attempts across 20 tests.
\end{itemize}
\begin{table*}[!t]
    \centering
    \renewcommand\arraystretch{1.2}
    \setlength\tabcolsep{2pt} 
    \setlength{\aboverulesep}{0pt}
    \setlength{\belowrulesep}{0pt} 
    \footnotesize 
    \definecolor{highlightblue}{HTML}{acbed2} 
    \begin{tabular}{cccccccccccc} 
        \hline
        \multirow{2}{*}{\textbf{Method}} & \multicolumn{5}{c}{\textit{\textbf{Cut Clay}}} & \multicolumn{2}{c}{\textit{\textbf{Put Bread into Pot}}} & \multicolumn{1}{c}{\textit{\textbf{Open Drawer}}} & \multicolumn{3}{c}{\textit{\textbf{Pour Balls}}} \\
        \cmidrule(lr){2-6}\cmidrule(lr){7-8}\cmidrule(lr){9-9}\cmidrule(l){10-12} % Adjusted cmidrule ranges
        & \textit{Pick (\%)} & \textit{Cut (\%)} & \textit{Sep. (\%)} & \textit{Place (\%)} & \textit{Attempt} $\downarrow$ & \textit{Succ. (\%)} $\uparrow$ & \textit{Attempt} $\downarrow$ & \textit{Succ. (\%)} $\uparrow$ & \textit{Pour (\%)} & \textit{Balls} $\uparrow$ & \textit{Pick (\%)} \\
        \hline
        %\cmidrule(r){1-1}\cmidrule{2-12}
        RISE~\cite{RISE} & 95 & 70 & 30 & 50 & \cellcolor{highlightblue}11 & 80 & 7 & 37.5 & 85 & 8.15 & 85 \\
        RISE w. \model & \cellcolor{highlightblue}100 & \cellcolor{highlightblue}90 & \cellcolor{highlightblue}55 & \cellcolor{highlightblue}65 & 12 & \cellcolor{highlightblue}95 & \cellcolor{highlightblue}1 & \cellcolor{highlightblue}52.5 & \cellcolor{highlightblue}100 & \cellcolor{highlightblue}9.60 & \cellcolor{highlightblue}100 \\
        \hline
        %\cmidrule(r){1-1}\cmidrule{2-12}
        \red{DP3~\cite{DP3}} & \red{\cellcolor{highlightblue}95} & \red{35} & \red{10} & \red{15} & \red{53} & \red{25} & \red{\cellcolor{highlightblue}16} & \red{20} & \red{55} & \red{1.20} & \red{60} \\
        \red{DP3 w. \model} & \red{80} & \red{\cellcolor{highlightblue}60} & \red{\cellcolor{highlightblue}20} & \red{\cellcolor{highlightblue}25} & \red{\cellcolor{highlightblue}25} & \red{\cellcolor{highlightblue}45} & \red{29} & \red{\cellcolor{highlightblue}55} & \red{\cellcolor{highlightblue}55} & \red{\cellcolor{highlightblue}1.65} & \red{\cellcolor{highlightblue}60} \\ 
        \hline
        %\cmidrule(r){1-1}\cmidrule{2-12}
        \red{DP~\cite{DP}} & \multicolumn{5}{c}{-} & \red{40} & \red{\cellcolor{highlightblue}12} & \red{20} & \red{10} & \red{0.80} & \red{30} \\  
        \red{DP w. \model} & \multicolumn{5}{c}{-} & \red{\cellcolor{highlightblue}45} & \red{16} & \red{\cellcolor{highlightblue}30} & \red{\cellcolor{highlightblue}40} & \red{\cellcolor{highlightblue}3.60} & \red{\cellcolor{highlightblue}60} \\ 
        \hline
    \end{tabular}
    \caption{\textbf{Performance comparison of policy models with and without MBA integration across four real-world tasks.}} 
    \label{real_result} 
\end{table*}
\red{We observe limitations with 2D policies on this long-horizon multi-object task as the appearance of clay is similar in 2D visual observations before and after the cutting process.}
%of clay in 2D visual observations before and after the cutting process.
%These 2D policies often directly move to the knife placement stage without executing any cutting actions.
\red{Therefore, we restrict our comparison to 3D policies.}
From Table.~\ref{real_result}, we observe that RISE with \model shows significant improvements over RISE in all stages of task execution, particularly in the cutting and separation stage.
\red{A similar improvement is observed with DP3.}
This improvement is attributed to the need for rotating the cutting surface and splitting it when the knife cuts into the clay, which is inherently a 6-DoF task.
\model leverages the prediction of the tool motion to provide feedback on the execution of the current action, resulting in superior task performance compared to the baseline.
However, \model still exhibits repeated attempts to grasp the knife.
We hypothesize that this issue arises from estimation errors in this task, where the thinness of the blade amplifies these errors.

\subsection{Put Bread into Pot}
% \begin{table}[!h]
% \centering
% \small
% \renewcommand\arraystretch{1.2}
% \begin{tabular}{c c c}
% \toprule
% \multirow{2}{*}{\textbf{Method}} & \multicolumn{2}{c}{\textbf{Metrics}} \\
% \cmidrule{2-3}
% \ & \textit{Success Rate} (\%) $\uparrow$ & \textit{Attempt} $\downarrow$ \\
% \midrule
% RISE & $80$ & $7$ \\
% RISE \textit{w.} \model &\cellcolor[HTML]{acbed2}$95$ &\cellcolor[HTML]{acbed2}$1$ \\
% \midrule
% DP3 & $25$ &\cellcolor[HTML]{acbed2}$16$ \\
% DP3 \textit{w.} \model &\cellcolor[HTML]{acbed2}$45$ &$29$ \\
% \midrule
% DP & $40$ &\cellcolor[HTML]{acbed2}$12$ \\
% DP \textit{w.} \model &\cellcolor[HTML]{acbed2}$45$ &$16$ \\
% \bottomrule
% \end{tabular}
% \caption{Experimental results of the \textbf{\textit{Put Bread into Pot}} task.}
% \label{realtable_bread}
% \end{table}

% \vspace{1em}
The \textbf{\textit{Put Bread into Pot}} task is a classic pick-and-place task, where the objective is to pick the bread from a bowl and place it into a pot.
We choose this task because bread is a soft object.
Unlike rigid bodies, soft objects can deform significantly due to the gripper's hold and the pressure from touching the bottom of the pot, which affects the object's pose, increasing the challenge for \model in predicting the object pose sequence.
We aim to assess the robustness and generalization ability of \model through this task.

In this experiment, we randomly initialize the positions of the pot and the bowl and change the orientation of the bread within the bowl. We record two evaluation metrics: the average success rate of placing the bread into the pot and the total number of redundant attempts to grasp the bread.

The results, as shown in Table.~\ref{real_result}, indicate that RISE with \model outperforms vanilla RISE by 15\% in average success rate. In most cases, RISE with \model successfully grasps the bread on the first attempt, whereas RISE often requires multiple attempts.
This demonstrates that \model can effectively handle object pose sequence prediction for soft objects, contributing to more accurate object localization and grasping. 

\red{Integrating \model with both DP and DP3 also notably improves success rates.
However, we also observe that a higher error rate in the base policy — primarily caused by the visual encoder — leads to increased attempts by \model.
This is likely due to the error accumulates when the visual backbone has a significant error, necessitating more attempts to compensate.}
% \begin{table*}[!t]
%     \centering
%     \renewcommand\arraystretch{1.2}
%     \setlength\tabcolsep{2pt}
%     \setlength{\aboverulesep}{0pt}
%     \setlength{\belowrulesep}{0pt}
%     \footnotesize
%     \begin{tabular}{c|ccccccccccc}
%         \hline
%         \multirow{2}{*}{\textbf{Method}}  & \multicolumn{2}{c}{\textit{\textbf{Put Bread into Pot}}} & \multicolumn{1}{c}{\textit{\textbf{Open Drawer}}} & \multicolumn{3}{c}{\textit{\textbf{Pour Balls}}} \\
%         \cmidrule(r){2-3}\cmidrule{4-4}\cmidrule(l){5-7}
%         & \textit{Success Rate} (\%) $\uparrow$ & \textit{Attempt} $\downarrow$ & \textit{Success Rate} (\%) $\uparrow$ & \textit{Pour}(\%) & \textit{Balls} $\uparrow$ & \textit{Pick}(\%) \\
%         \hline
%         DP \textit{w.} ATM & $40$ &\cellcolor[HTML]{acbed2}$2$ & $5$ & $25$ & $2.05$ & $35$  \\
%         DP \textit{w.} \model & \cellcolor[HTML]{acbed2}$45$ &  $16$ &\cellcolor[HTML]{acbed2}$30$ &\cellcolor[HTML]{acbed2}$40$ & \cellcolor[HTML]{acbed2}$3.60$ & \cellcolor[HTML]{acbed2}$60$ \\
%         \hline
%     \end{tabular}
%     \caption{\textbf{Comparison of policy models under different object motion conditions}: \textbf{\model (ours)}: 6 DoF trajectory \textit{vs.} \textbf{ATM} \cite{ATM}: point tracks}
%     \label{vs_flow}
% \end{table*}

\subsection{Open Drawer}
The \textbf{\textit{Open Drawer}} task is a two-stage process where the robot must first precisely grasp the drawer handle and then pull it open horizontally.
The main difficulty lies in the small distance between the handle and the drawer surface, where even slight positional errors can result in a failed grasp or cause the gripper to lose contact during the pulling motion.
This task challenges the policy's ability to accurately estimate the handle pose sequence to ensure smooth and high-precision execution.

In this experiment, we account for the fact that delicate operations can be easily influenced by scene setup, particularly issues like point cloud occlusion when the robotic arm manipulates drawers at low heights.
To minimize the impact of randomness, we double the number of test trials from the original plan, increasing it to 40 trials.
Our test positions cover a variety of positions and orientations in the workspace, including areas within both the inner circle and the outer ring of the workspace plane.
Each area receives 20 test trials.
We also include scenarios where the initial drawer position causes point cloud occlusion, ensuring a comprehensive evaluation of the policies' robustness.
% We report the average success rates of the two task stages (\textit{Grasp}: grasp the handle; \textit{Pull}: pull the drawer until open) as evaluation metrics to accurately assess the execution capabilities of the policies.
We report the average success rates over all the trials.
% \begin{table}[!h]
% \centering
% \renewcommand\arraystretch{1.2}
% \small
% \begin{tabular}{c c}
% \toprule
% \textbf{Method} & \textbf{Success Rates} (\%) $\uparrow$ \\
% \midrule
% RISE & $37.5$ \\
% RISE \textit{w.} \model & \cellcolor[HTML]{acbed2}$52.5$ \\
% \midrule
% DP3 & $20$ \\
% DP3 \textit{w.} \model & \cellcolor[HTML]{acbed2}$55$ \\
% \midrule
% DP & $20$ \\
% DP \textit{w.} \model & \cellcolor[HTML]{acbed2}$30$ \\
% \bottomrule
% \end{tabular}
% \caption{Experimental Results of the \textbf{\textit{Open Drawer}} task.}
% \label{realtable_drawer}
% \end{table}

The experimental results are shown in Table.~\ref{real_result} indicate that policies with \model outperform the baselines in both stages of the task. This aligns with our findings in simulation experiments, further confirming that integrating \model significantly enhances the policies in tasks that require precise object localization and fine manipulation.
% \subsection{Pour Balls}
% \begin{table}[!h]
% \centering
% \renewcommand\arraystretch{1.2}
% \small
% \begin{tabular}{c c c c c}
% \toprule
% \multirow{2}{*}{\textbf{Method}} & \multicolumn{3}{c}{\textbf{Metrics}} \\
% \cmidrule{2-4}
% \ & \textit{Pour}(\%) & \textit{Balls} $\uparrow$ & \textit{Pick}(\%) \\
% \midrule
% RISE & $85$ & $8.15$ &$85$\\
% RISE \textit{w.} \model &\cellcolor[HTML]{acbed2}$100$ &\cellcolor[HTML]{acbed2}$9.60$&\cellcolor[HTML]{acbed2}$100$ \\
% \midrule
% DP3 & $55$ & $1.20$ &$60$\\
% DP3 \textit{w.} \model &\cellcolor[HTML]{acbed2}$55$ &\cellcolor[HTML]{acbed2}$1.65$&\cellcolor[HTML]{acbed2}$60$ \\
% \midrule
% DP & $10$ & $0.80$ &$30$\\
% DP \textit{w.} \model &\cellcolor[HTML]{acbed2}$40$ &\cellcolor[HTML]{acbed2}$3.60$&\cellcolor[HTML]{acbed2}$60$ \\
% \bottomrule
% \end{tabular}
% \caption{Experimental Results of the \textbf{\textit{Pour Balls}} task.}
% \label{realtable_balls}
% \end{table}
% \vspace{1em}
\subsection{Pour Balls}
The \textbf{\textit{Pour Balls}} task is a challenging one, where the objective is to lift a cup filled with 10 balls and pour them into a bowl. The difficulty arises from two main factors:

\begin{enumerate}[label=\textbf{\arabic*)},leftmargin=12pt]
    \item The cup is a non-cubic object, with varying width at different heights. Therefore, the gripper must learn a precise visual-to-motion control policy to grasp the cup at the correct height. If the gripper is too wide, it will fail to hold the cup, causing it to fall; if the gripper is too narrow, it will knock over the cup.
    \item This is a 6-DoF task, and during the pouring process, if the translation and rotation are not properly controlled, or the gripper's force is not correctly adjusted, the cup will either rotate or fall, resulting in task failure.
\end{enumerate}

In this round of experiments, we still randomize the object positions. 
We conduct 20 trials.
In this experiment, we consider three evaluation metrics: the average success rate of pouring balls into the bowl, the average number of balls poured into the bowl across all trials, and the average success rate of picking up the cup.

In this experiment, \model achieves more than 15\% the success rate compared to RISE \red{and 30\% compared to DP}, as shown in Table \ref{real_result}.
Notably, under the same successful conditions of picking up the cup, \model is also able to pour the balls with greater precision.
This indicates that \model can accurately capture the relationship between object pose variations and corresponding actions in 6-DoF tasks, making it well-suited for handling intricate manipulation scenarios.

\red{\subsection{Comparative Experiment to Flow-based methods}}
\red{In this section, we compare \model against the representative flow-based method ATM~\cite{ATM} to demonstrate that the pose prediction paradigm yields higher quality action generation for the policy compared to prediction in the visual space.
Specifically, both ATM and \model employ a diffusion action head for prediction; \model conditions on predicted future object pose sequences, whereas ATM conditions on keypoint flow predicted by a Track Transformer.}

\red{We evaluate \model on three 2D policy-fit tasks and found that it outperforms ATM across most metrics, as shown in Table~\ref{vs_flow}.
Notably, in tasks requiring fine-grained manipulation, such as \textbf{\textit{Open Drawer}}, the flow-based approach's tracking is confined to a limited number of pixels in the visual space, making it challenging to capture the pose of the handle, which can lead to grasping failures.
In contrast, tasks with higher degrees of freedom, like \textbf{\textit{Pour Balls}}, necessitate more comprehensive spatial motion modeling and precise control over the object's rotational dynamics.
In these scenarios, flow-based policies appear to be at a disadvantage.
The results validate our hypothesis regarding the ``vision-motion'' gap, underscoring the necessity of modeling pose information.}

\begin{table}[!t]
    \centering
    \renewcommand\arraystretch{1.2}
    \setlength\tabcolsep{4pt}
    \setlength{\aboverulesep}{0pt}
    \setlength{\belowrulesep}{0pt}
    \footnotesize
    \begin{tabular}{cccc}
        \hline
        \red{\textbf{Task}} & \red{\textbf{Metric}} & \red{DP \textit{w.} ATM~\cite{ATM}} & \red{DP \textit{w.} \model} \\
        \hline
        \multirow{2}{*}{\shortstack[c]{\red{\textit{\textbf{Put Bread}}}\\\red{\textit{\textbf{into Pot}}}}} 
            & \red{\textit{Success} (\%) $\uparrow$} & \red{$40$} & \red{\cellcolor[HTML]{acbed2}$45$} \\
            & \red{\textit{Attempt} $\downarrow$} & \red{\cellcolor[HTML]{acbed2}$2$} & \red{$16$} \\
        \hline
        \multirow{1}{*}{\red{\textit{\textbf{Open Drawer}}}} 
            & \red{\textit{Success} (\%) $\uparrow$} & \red{$5$} & \red{\cellcolor[HTML]{acbed2}$30$} \\
        \hline
        \multirow{3}{*}{\red{\textit{\textbf{Pour Balls}}}} 
            & \red{\textit{Pour} (\%)} & \red{$25$} & \red{\cellcolor[HTML]{acbed2}$40$} \\
            & \red{\textit{Balls} $\uparrow$} & \red{$2.05$} & \red{\cellcolor[HTML]{acbed2}$3.60$} \\
            & \red{\textit{Pick} (\%)} & \red{$35$} & \red{\cellcolor[HTML]{acbed2}$60$} \\
        \hline
    \end{tabular}
    \caption[]{\red{\textbf{Comparison of policy models under different object motion conditions}}: \red{\textbf{\model (ours)}}: \red{$6$D poses} \red{\textit{vs.}} \red{\textbf{ATM}}: \red{2D point flow.}}
    \label{vs_flow}
\end{table}

\red{\subsection{Inference Speed}}
\red{We evaluated the average inference time (excluding visual backbone processing time for experimental consistency across different backbones) for DP, ATM, and DP with \model. DP achieved 95.98 ms, ATM 105.85 ms, while DP with \model required 197.50 ms. This demonstrates the increased computational cost associated with \model's precise control. ATM benefits from its transformer-based flow tracking, avoiding computationally expensive denoising steps.}

\sectionred{Limitations}
\red{\model still exhibits several limitations. 
First, its inference efficiency can be improved. Future work might focus on replacing the existing denoising paradigm with more computationally efficient models~\cite{ACT,su2025dense}. 
Second, \model's object pose supervision relies on ground-truth data from a MoCap system, which incurs high acquisition costs. Borrowing from flow-based methods that utilize flow generative models to provide ground truth~\cite{ATM,YuanGeneralFlow} offers a promising alternative. Correspondingly, leveraging foundation 6D pose estimation models~\cite{wen2024foundationpose,hirschorn2023pose} for data annotation is also a valuable direction.
Third, \model is limited in handling varying numbers of objects, as its predefined observation vector constrains it to a fixed object count. 
Finally, the manipulation of deformable objects, whose 6D pose is not trackable, remains unaddressed. 
These limitations point to important directions for future research.}
% Finally, \model's current object manipulation capabilities are limited. The predefined observation vector restricts its applicability to a fixed number of objects, and manipulation of more complex deformable objects remains unaddressed in this work. These are important directions for future research. 

\section{Conclusion}
In this paper, we introduce \model, a novel module that draws inspiration from the reasoning process of human beings. \model infers future object motion sequences and uses them as guidance for action generation. It can be flexibly integrated into existing robotic manipulation policies with diffusion action heads in a plug-and-play manner, significantly improving the performance of these policies across a wide range of tasks. This work holds great potential for future development, including integrating it into other robotic manipulation policies with different action heads, utilizing diverse object motion demonstration data (e.g., human demonstrations or web videos) for supervision and learning, exploring its performance in long-horizon, multi-stage tasks for both object pose and action prediction, and expanding it into a general large-scale policy across multi-dataset, multi-task settings.

\renewcommand*{\bibfont}{\footnotesize}
\printbibliography

\end{document}